\documentclass[journal]{IEEEtran}

\usepackage{graphicx}
\usepackage{comment}
\usepackage{amsmath,amssymb}
\usepackage{color}
\usepackage{booktabs}

\usepackage{xspace}
\usepackage{hyperref}

\makeatletter
\DeclareRobustCommand\onedot{\futurelet\@let@token\@onedot}
\def\@onedot{\ifx\@let@token.\else.\null\fi\xspace}
\def\eg{\emph{e.g}\onedot} 
\def\ie{\emph{i.e}\onedot} 
 
\def\etc{\emph{etc}\onedot} \def\vs{\emph{vs}\onedot}
 
\def\etal{\emph{et al}\onedot}
\makeatother

\usepackage{array}
\newcolumntype{C}[1]{>{\centering\arraybackslash}p{#1}}
\usepackage{multirow}

\begin{document}

\title{DTVNet+: A High-Resolution Scenic Dataset for \\ Dynamic Time-lapse Video Generation}

\author{Jiangning~Zhang,
        Chao~Xu,
        Yong~Liu,
        and~Yunliang~Jiang
\thanks{This work extends the previous DTVNet in ECCV'20, and we further present a high-quality and high-resolution Quick-Sky-Time dataset and carry out more adequate experiments and analysis.}
\thanks{Code is available at \url{https://github.com/zhangzjn/DTVNet}.}
}

\maketitle

\begin{abstract}
  This paper presents a novel end-to-end dynamic time-lapse video generation framework, named DTVNet, to generate diversified time-lapse videos from a single landscape image conditioned on normalized motion vectors. The proposed DTVNet consists of two submodules: \emph{Optical Flow Encoder} (OFE) and \emph{Dynamic Video Generator} (DVG). The OFE maps a sequence of optical flow maps to a \emph{normalized motion vector} that encodes the motion information of the generated video. The DVG contains motion and content streams to learn from the motion vector and the single landscape image. Besides, it contains an encoder to learn shared content features and a decoder to construct video frames with corresponding motion. Specifically, the \emph{motion stream} introduces multiple \emph{adaptive instance normalization} (AdaIN) layers to integrate multi-level motion information for controlling the object motion. In the testing stage, videos with the same content but various motion information can be generated by different \emph{normalized motion vectors} based on only one input image. Also, we propose a high-resolution scenic time-lapse video dataset, named Quick-Sky-Time, to evaluate different approaches, which can be viewed as a new benchmark for high-quality scenic image and video generation tasks. We further conduct experiments on Sky Time-lapse, Beach, and Quick-Sky-Time datasets. The results demonstrate the superiority of our approach over state-of-the-art methods for generating high-quality and various dynamic videos.
\end{abstract}


\IEEEpeerreviewmaketitle

\section{Introduction} \label{Intro}
Video generation is a task to generate video sequences from the noise or special conditions such as images and masks, which has promising application capabilities, \eg, video dataset expansion, texture material generation, and film production. However, this is a challenging task where the model has to learn the content, motion, and relationship between objects simultaneously, especially when objects have no special shapes such as cloud and fog. We focus on a more challenging task in this work that generates realistic and dynamic time-lapse videos based on a natural landscape image.

Thanks to the development of \emph{generative adversarial network}~\cite{gan,WGAN,IWGAN}, many excellent models have been proposed to generate realistic and dynamic videos~\cite{MDGAN,Mocogan,chen2017video,zhao2018learning,dong2019fw,wang2020imaginator,zhang2020dtvnet}.
Vondrick~\etal~\cite{VGAN} explore how to leverage a random noise to learn the dynamic video in line with a moving foreground and static background pathways from large amounts of unlabeled video by capitalizing on adversarial learning methods. However, this kind of noise-based method usually suffers from low-quality video generation and hard training because of its mapping from a vector to a high-resolution feature map.
Later work~\cite{villegas2017learning} adopts the LSTM~\cite{LSTM} structure to predict future high-level poses in a sequence to sequence manner and then use an analogy-based encoder-decoder model to generate future images. The method can generate high-quality and reasonable images, but it acquires extra-label information such as human pose and several past reference images that are unpractical.
Subsequent MoCoGAN~\cite{Mocogan} apply the LSTM model to generate a sequence of random vectors that consist of content and motion parts and then map them to a sequence of video frames. This method can generate diversified dynamic videos that contain the same content in theory because we can fix the content part while changing the motion part of the random vectors. Nevertheless, the resolution of the generated video is still low.

To obtain high-resolution and high-quality time-lapse videos, some researchers propose using spatio-temporal 3d convolutions in all encoder and decoder modules, aiming to capture both the spatial and temporal components of a video sequence. Nevertheless, this kind of method is time-consuming during the training stage for utilizing concepts of the progressively growing GAN~\cite{ProGAN}. Xiong~\etal~\cite{MDGAN} recently propose a two-stage 3DGAN-based model named MDGAN, which learns to generate long-term realistic time-lapse videos of high resolution given the first frame. Though MDGAN is capable of generating vivid motion and realistic video, it is hard to train such a two-stage model and lacks diverse video generation capabilities, limiting the practicality of the method. We follow the same time-lapse video generation task with MDGAN and design a practical one-stage model.

\begin{figure*}[t]
    \centering
    \includegraphics[width=1.0\textwidth]{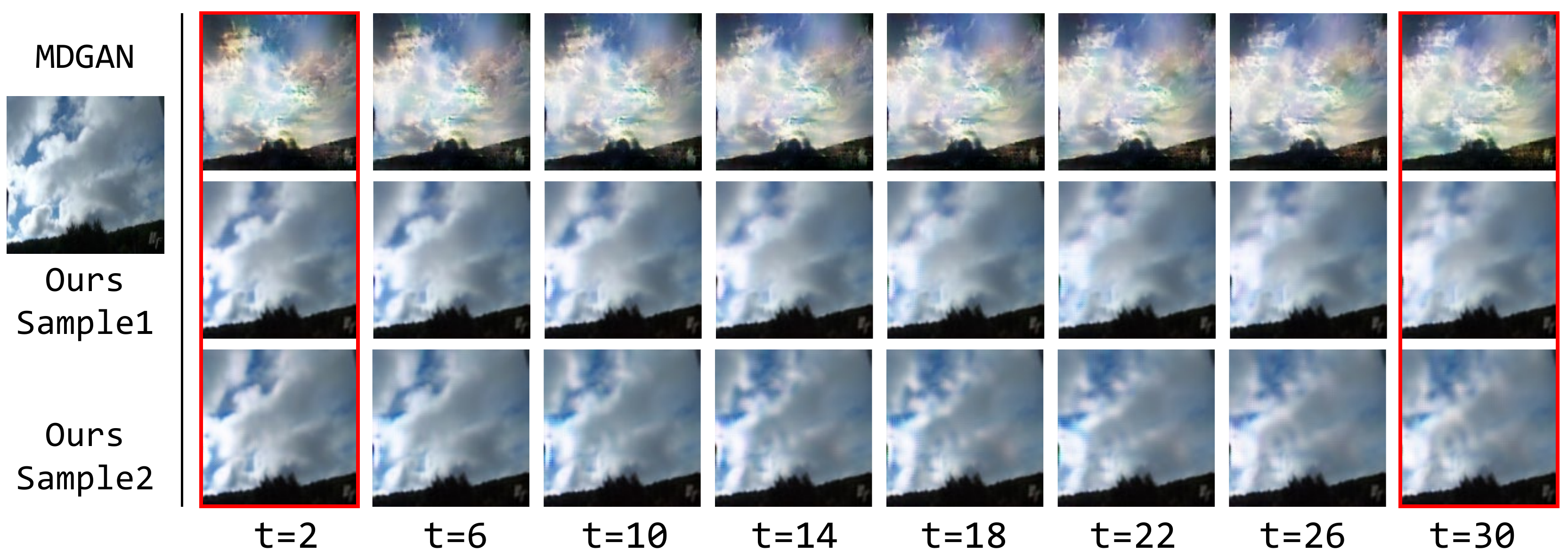}
    \caption{The first row shows generated frames by the state-of-the-art MDGAN~\cite{MDGAN}, and the other two rows are from our method with random motion vectors. The left image is in-the-wild start landscape frame, and other frames are generation results at different times. Our method could generate more photorealistic and diversified video frames. Please zoom in the red rectangles for a more precise comparison.}
    \label{fig:comparison1}
    \vspace{-0.5em}
\end{figure*}

Some researchers recently introduce optical flow maps that contain the motion information during the training and testing stages, aiming at explicitly supplying motion signals to the network. Liang~\etal~\cite{flow1} design a dual motion GAN that simultaneously solves the primal future-frame prediction and future-flow prediction tasks. Though it can generate high-quality video frames, it is time-consuming to take more computational effort on future-flow predictions.
Besides the quality and the resolution, the diversification of the generated time-lapse videos is crucial for practical application. Li~\etal~\cite{flow2} first maps a sampled noise to consecutive flows and then uses the proposed video prediction algorithm to synthesize a set of likely future frames in multiple time steps from one single still image. During the testing, the method can directly use sample points from the distribution for predictions. Unlike the aforementioned methods, our method is designed in an end-to-end manner. We introduce a normalized motion vector to control the generation process, which can generate high-quality and diversified time-lapse videos.

In this work, we propose an end-to-end dynamic time-lapse video generation framework, \ie, DTVNet, to generate diversified time-lapse videos from a single landscape image. The time-lapse landscape videos generally contain still objects, \eg, house, earth, and tree, as well as unspecific objects, \eg, cloud, and fog, which is challenging to understand the motion relationship between objects.
Specifically, the proposed DTVNet consists of two submodules: \emph{Optical Flow Encoder} (OFE) and \emph{Dynamic Video Generator} (DVG). OFE introduces an unsupervised optical flow estimation method to get the motion maps among consecutive images and encoders them to a normalized motion vector. DVG contains motion and content streams to learn from the motion vector and the single landscape image. Besides, it contains an encoder to learn shared content features and a decoder to construct video frames with corresponding motion. In detail, the normalized motion vector is integrated into the motion stream by multiple AdaIN layers~\cite{AdaIN}. We apply content loss, motion loss, and adversarial loss during the training stage to ensure high-quality, dynamic, and diversified video generation, as shown in Figure~\ref{fig:comparison1}. During the testing stage, we exclude the process of flow estimation and flow encoding and directly sample from a normalization distribution as the motion vector, which reduces the network computing overhead and supplies diversified motion information simultaneously.

Specifically, we make the following five contributions:
\begin{itemize}
\item An \emph{optical flow encoder} (OFE) is designed to supply normalized motion information in the training stage that is used to guide diversified video frame generation.
\item A new \emph{dynamic video generator} (DVG) is proposed first to learn disentangling content and motion features separately and then use integrated features to generate target video.
\item We apply content loss, motion loss, and adversarial loss during the training stage, ensuring high-quality, dynamic, and diversified video generation.
\item We build a large-scale and high-resolution Quick-Sky-Time (QST) dataset to evaluate different approaches, which can also be viewed as a new benchmark for generating high-quality videos and images.
\item Experimental results on Sky Time-lapse, Beach, and QST datasets demonstrate the superiority of our approach, which can generate high-quality and dynamic video frames in an end-to-end one-stage network.
\end{itemize}

\section{Related Work} \label{Related}

\noindent\textbf{Optical Flow Estimation.}
Optical flow is a reliable representation to characterize motion between frames. Starting from Flownet~\cite{FlowNet}, many supervised methods for optical flow estimation are proposed, \eg FlowNet2~\cite{FlowNet2}, PWC-Net~\cite{PWCNet}, IRR-PWC~\cite{IRR-PWC}, \etc. Though these methods are in high accuracy and efficiency, they heavily depend on the labeled dataset. It is hard to acquire the ground truth in reality, which reduces the practicality of these methods.

As an alternative, some researchers focus on studying unsupervised methods~\cite{yin2018geonet,wang2018occlusion,jason2016back,CCFlow,EpiFlow} and have achieved great success. Liu ~\etal~\cite{SelFlow} propose the SelFlow that distills reliable flow estimations from non-occluded pixels and uses these predictions as
ground truth to learn optical flow for hallucinated occlusions. DDFlow~\cite{DDFlow} further improves the model performance by distilling unlabeled data. In this paper, we first apply the unsupervised method~\cite{arflow} to estimate the optical flow map and then encoder the flow information to a motion vector that is used as a condition to guide the video generation procedure.

\noindent\textbf{Generative Adversarial Networks.}
Since Goodfellow~\etal~\cite{gan} first introduces the generative adversarial network (GAN) that contains a generator and a discriminator, many GAN-based approaches are proposed~\cite{WGAN,IWGAN,wu2016learning,mao2017least,nowozin2016f,qi2020loss,salimans2016improved,stylegan2,msggan,gan_review}. They have achieved impressive applications in various fields~\cite{freenet,esrgan,spsr,tmm1,tmm2,tmm3,MDGAN,Mocogan}, \eg, facial manipulation, super-resolution, video understanding and generation, \etc. Mehdi~\etal~\cite{cGAN} propose the cGAN that controls the mode of generated samples by adding an extra conditional variable to the network. Pix2Pix~\cite{Pix2Pix} uses $\ell_1$ and adversarial loss for paired image translation tasks, and Zhu~\etal~\cite{cycleGAN} further introduces a cycle consistency loss to deal with unpaired image-to-image translation tasks. ProGAN~\cite{ProGAN} describes a new training methodology that grows both the generator and discriminator progressively, capable of generating up to 1,024 resolution images. Besides 2D-based GAN methods, Wu~\etal~\cite{wu2016learning} apply 3D convolution to generate 3D objects from a probabilistic space. MDGAN~\cite{MDGAN} presents a 3D convolutional-based two-stage approach to generate realistic time-lapse videos of high resolution. Our model follows the GAN idea and considers extra motion information when generating videos.

\noindent\textbf{Video Generation.}
Video generation aims at generating image sequences from a noise, image(s), or with an extra condition such as human pose, semantic label map, and optical flow~\cite{balaji2019conditional,li2018video,fan2019controllable,Pose,ohnishi2018hierarchical}.
Mathieu~\etal~\cite{mathieu2015deep} first adopt the GAN idea to mitigate the inherently blurry predictions obtained from the standard mean squared error loss function. Subsequently, VGAN~\cite{VGAN} is proposed to untangle the foreground from the background of the scene with a spatio-temporal convolutional architecture, and many follow-up works borrowed the idea of disentangling. Saito~\etal~\cite{TemGAN} exploits two different types of generators, \ie, a temporal generator and an image generator, to generate videos and achieve good performance. At the same time, MoCoGAN~\cite{Mocogan} maps a sequence of random vectors that consists of content and motion parts to a sequence of video frames. 
To obtain high-resolution and high-quality videos, Aigner~\etal~\cite{FutureGAN} propose the FutureGAN that utilizes concepts of the progressively growing GAN~\cite{ProGAN} and adopts spatio-temporal 3d convolutions to capture both the spatial and temporal components of a video sequence.  

However, these noise-input and progressively growing methods generally suffer from a low-quality generation or a complicated training process, so the following methods that use a single image as input are proposed. MDGAN~\cite{MDGAN} adopts a two-stage network to generate long-term future frames. It generates videos of realistic content for each frame in the first stage and then refines the generated video from the first stage. Nam~\etal~\cite{nam2019end} learns the correlation between the illumination change of an outdoor scene and the time of the day by a multi-frame joint conditional generation framework. Yang~\etal~\cite{Pose} propose a pose-guided method to synthesize human videos in a disentangled way: plausible motion prediction and coherent appearance generation. Similarly, Cai~\etal~\cite{Pose1} design a skeleton-to-image network to generate human action videos. Researches~\cite{vid2vid1,vid2vid2} take one semantic label map as input to synthesize a sequence of photo-realistic video frames. Recently, some flow-based methods have made a great success. Liang~\etal~\cite{flow1} design a dual motion GAN that simultaneously solves the primal future-frame prediction and future-flow prediction tasks. Li~\etal~\cite{flow2} propose a video prediction algorithm that synthesizes a set of likely future frames in multiple time steps from one still image.
Considering the resolution and the motion of generated frames, we design our model as two streams: motion and content streams for solving the motion vector and the image information, respectively, and then fuse features from both streams to generate time-lapse video frames. Our model neither has to map random vectors to a sequence of video frames from scratch nor generates video sequences frame by frame.

\begin{figure*}[t]
    \centering
    \includegraphics[width=1.0\linewidth]{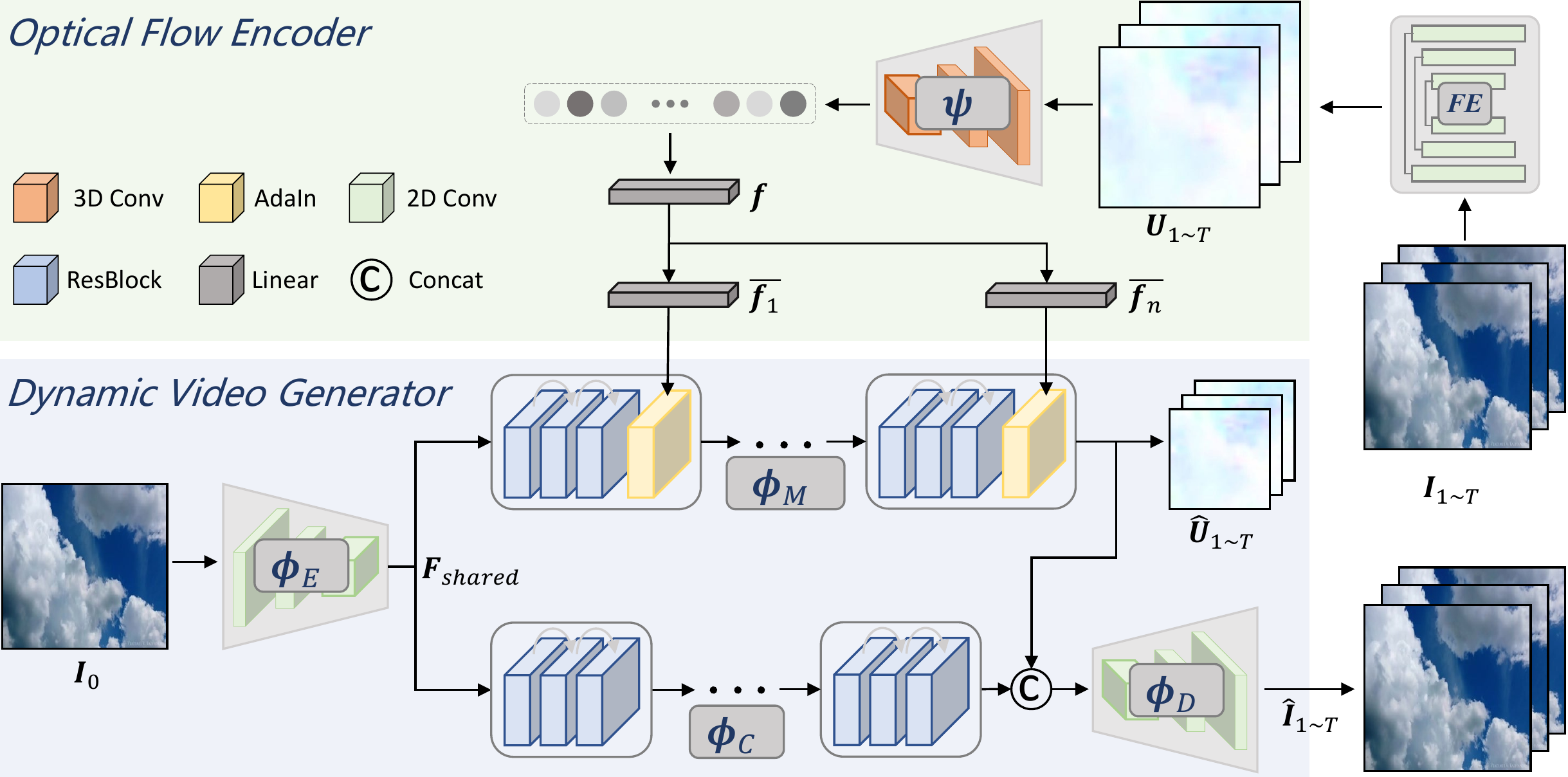}
    \caption{Overview of the proposed DTVNet that consists of a \emph{Optical Flow Encoder} $\boldsymbol{\psi}$ and a \emph{Dynamic Video Generator} $\boldsymbol{\phi}$. Given the first landscape image $\boldsymbol{I}_0$ and subsequent landscape images $\boldsymbol{I}_{1\sim32}$, the \emph{flow estimator} ($\boldsymbol{FE}$) first estimates the consecutive flows $\boldsymbol{U}_{1\sim T}$ and then $\boldsymbol{\psi}$ encodes the flows to the \emph{normalized motion vector} $\bar{\boldsymbol{f}}$. Dynamic video predictor $\boldsymbol{\phi}$ successively apply encoder $\boldsymbol{\phi}_{E}$ to learn shared content feature, motion stream $\boldsymbol{\phi}_{M}$ and content streams $\boldsymbol{\phi}_{C}$ to learn motion and content information from $\bar{\boldsymbol{f}}$ and $\boldsymbol{I}_{0}$, respectively, and a decoder $\boldsymbol{\phi}_{D}$ to construct consecutive video frames $\hat{\boldsymbol{I}}_{1\sim32}$.}
    \label{fig:DTVNet}
\end{figure*}

\section{Method} \label{Method}
In this paper, a novel end-to-end dynamic time-lapse video generation framework named DTVNet is proposed to generate diversified time-lapse videos from a single landscape image. Formally, given a single landscape image $\boldsymbol{I}_{0}$, the model learns to generate a time-lapse video sequences $\hat{\boldsymbol{V}} = \{\hat{\boldsymbol{I}}_{1}, \hat{\boldsymbol{I}}_{2}, \dots, \hat{\boldsymbol{I}}_{T}\}$. As depicted in Figure~\ref{fig:DTVNet}, DTVNet consists of \emph{Optical Flow Encoder} ($\boldsymbol{\psi}$) and \emph{Dynamic Video Generator} ($\boldsymbol{\phi}$), and we will explain our approach as follows. Note the notations: thin lower and upper cases for scalars, bold lower case for vectors, and bold upper case for matrices.

\subsection{Optical Flow Encoder} \label{OFE}

Considering the motion and diversity of the generated video, we design an \emph{optical flow encoder} (OFE) module to encode the motion information to a \emph{normalized motion vector}, thus not only can it supply motion information but also we could sample various motion vectors from a normalization distribution to generate diversified videos in the testing stage.

As shown in Figure~\ref{fig:DTVNet}, we first apply unsupervised optical flow estimator which is more practical for not requiring label information (we use SelFlow~\cite{SelFlow} in this paper) to estimate consecutive flows $\boldsymbol{U}_{1\sim T} = \left \{\boldsymbol{U}_{0\rightarrow1}, \boldsymbol{U}_{1\rightarrow2}, \dots, \boldsymbol{U}_{(T-1)\rightarrow T}\right\}$ from real video frames $\boldsymbol{V}_{0\sim T} = \left\{\boldsymbol{I}_{0}, \boldsymbol{I}_{1}, \dots, \boldsymbol{I}_{T}\right\}$, where $T$ indicates maximum frame. We formulate this process as:
\begin{align}
  \boldsymbol{U}_{1\sim T} = \boldsymbol{FE}(\boldsymbol{V}_{0\sim T}).
\end{align}

Then OFE module $\boldsymbol{\psi}$ encode consecutive flows $\boldsymbol{U}_{1\sim T}$ to a normalized motion vector $\boldsymbol{f}$ that contains the motion information, denoted as:
\begin{align}
  \boldsymbol{f} = \boldsymbol{\psi}(\boldsymbol{U}_{1\sim T}).
\end{align}

Specifically, OFE employs a 3D encoder architecture, which is proven to be more suitable for learning spatial-temporal features than 2D convolution~\cite{C3D}. With the reduction of the time dimension, the 3D encoder can model the motion information of local adjacent frames and the global motion information of the sequence.

As a result, the normalized 512-dimensional vector that contains continuous motion information is integrated into the DVG module. Detailed structure and parameters can be found in the supplementary material.

\subsection{Dynamic Video Generator} \label{DVG}
To generate photo-realistic and vivid video that has consistent content with the reference image and dynamic movement, we propose a novel \emph{dynamic video generator} (DVG) that is designed in a disentangling idea. As shown in Figure~\ref{fig:DTVNet}, DVG contains an encoder $\boldsymbol{\phi}_{E}$, a motion stream $\boldsymbol{\phi}_{M}$, a content streams $\boldsymbol{\phi}_{C}$, and a decoder $\boldsymbol{\phi}_{D}$.

In detail, the encoder $\boldsymbol{\phi}_{E}$ learns shared content feature $\boldsymbol{F}_{shared}$ from the first landscape image $\boldsymbol{I}_{0}$, denoted as:
\begin{align}
  \boldsymbol{F}_{shared} = {\boldsymbol{\phi}_{E}}(\boldsymbol{I}_{0}).
\end{align}

Considering the mismatch between the normalized motion vector $f$ and motion stream features in semantic level that could inevitably increase the training difficulty, we introduce multiple linear layers to learn the adaptive motion vectors $\left\{\bar{\boldsymbol{f}}_{1}, \dots, \bar{\boldsymbol{f}}_{n}\right\}$. Then adaptive motion vectors are integrated into the motion stream ${\boldsymbol{\phi}_{M}}$ by multiple AdaIN layers. Specifically, adaptive motion vector $\bar{\boldsymbol{f}}_{i}$ is first specialized to motion styles $\boldsymbol{z}_{i}=(\boldsymbol{z}_{i}^{scale}, \boldsymbol{z}_{i}^{shift})$ for $i_{th}$ AdaIN layer with the input feature map $\boldsymbol{F}_{i}^{in}$. Then we can calculate the output feature map $\boldsymbol{F}_{i}^{out}$ in the following formula:
\begin{align}
  \boldsymbol{F}_{i}^{out} = \boldsymbol{z}_{i}^{scale} \frac{\boldsymbol{F}_{i}^{in} - \mu (\boldsymbol{F}_{i}^{in})}{\sigma (\boldsymbol{F}_{i}^{in})} + \boldsymbol{z}_{i}^{shift},
\end{align}
where $\mu(\cdot)$ and $\sigma(\cdot)$ calculate mean and variance, respectively. Complete formula for motion stream is as follows:
\begin{align}
    {\hat{\boldsymbol{U}}}_{1\sim T} = {\boldsymbol{\phi}_{M}}(\boldsymbol{F}_{shared}, \bar{\boldsymbol{f}}_{1}, \dots, \bar{\boldsymbol{f}}_{n}),
\end{align}
where ${\hat{\boldsymbol{U}}}_{1\sim T}$ indicates adapted low-resolution flows. During the training stage, ${\hat{\boldsymbol{U}}}_{1\sim T}$ is supervised by real optical flows, aiming at adapting the motion to the input landscape image.

Analogously, the content stream $\boldsymbol{\phi}_{C}$ also use $\boldsymbol{F}_{shared}$ as input to further learn deeper features. Subsequent decoder $\boldsymbol{\phi}_{D}$ synthesize target video by combined motion and content information:
\begin{align}
    {\hat{\boldsymbol{I}}}_{1\sim T} = {\boldsymbol{\phi}_{D}}({\hat{\boldsymbol{U}}}_{1\sim T}, {\boldsymbol{\phi}_{C}}(\boldsymbol{F}_{shared})),
\end{align}

\subsection{Objective Function}
During the training stage of the DTVNet, we adopt content loss to monitor image quality at the pixel level, motion loss to ensure reasonable movements of the generated video, and adversarial loss to further boost video quality and authenticity. The full loss function $\mathcal{L}_{all}$ is defined as follow:
\begin{align}
    \mathcal{L}_{all} = \lambda_{C} \mathcal{L}_{C} + \lambda_{M} \mathcal{L}_{M} + \lambda_{adv} \mathcal{L}_{adv},
\end{align}
where $\lambda_{C}$, $\lambda_{M}$, and $\lambda_{adv}$ represent weight parameters to balance different terms.

\textbf{Content Loss.} The first term $\mathcal{L}_{C}$ calculates $\ell_1$ errors between generated images ${\hat{\boldsymbol{I}}}_{1\sim T}$ and real images $\boldsymbol{I}_{1\sim T}$.
\begin{align}
    \mathcal{L}_{C} &= \sum_{i=1}^{T} || {\hat{\boldsymbol{I}}}_{i} - \boldsymbol{I}_{i} ||_{1}.
\end{align}

\textbf{Motion Loss.} The second term $\mathcal{L}_{M}$ calculates $\ell_1$ errors between adapted low-resolution flows ${\hat{\boldsymbol{U}}}_{1\sim T}$ and real optical flows ${\boldsymbol{U}}_{1\sim T}^{LR}$. Note that $\boldsymbol{U}_{1\sim T}^{LR}$ are reconstructed low-resolution optical flow maps from $\boldsymbol{U}_{1\sim T}$.
\begin{align}
    \mathcal{L}_{M} &= \sum_{i=1}^{T} || {\hat{\boldsymbol{U}}}_{i} - \boldsymbol{U}_{i}^{LR} ||_{1}.
\end{align}

\textbf{Adversarial Loss.} The third term $\mathcal{L}_{adv}$ employ the improved WGAN with a gradient penalty for adversarial training~\cite{IWGAN}. Specifically, the discriminator $D$ consists of six (3D-Conv)-(3D-InNorm)-(LeakyReLU) blocks that can capture discriminative spatial and temporal features.
\begin{align}
    \mathcal{L}_{GAN} = & \mathbb{E}_{{\tilde{\boldsymbol{V}}} \sim p_{g}}[D({\tilde{\boldsymbol{V}}})] - \mathbb{E}_{\boldsymbol{V} \sim p_{r}}[D(\boldsymbol{V})] + \notag \\
    &\lambda \mathbb{E}_{{\hat{\boldsymbol{V}}} \sim p_{{\hat{\boldsymbol{V}}}}}[(||\nabla_{{\hat{\boldsymbol{V}}}}D({\hat{\boldsymbol{V}}})||_2 - 1) ^ 2],
\end{align}
where $p_{r}$ and $p_{g}$ are real and generated video distribution, respectively, and $p_{\hat{x}}$ is implicitly defined by sampling uniformly along straight lines between pairs of points sampled from $p_{r}$ and $p_{g}$.

\subsection{Training Scheme} \label{TS}
We first train the unsupervised optical flow estimator (FE) under the instruction of SelFlow~\cite{SelFlow} and fix its parameters once the training is complete in all experiments. When training the DTVNet, we set loss weights $\lambda_{C}$, $\lambda_{M}$, and $\lambda_{adv}$ to 100, 1, and 1, respectively. The layer number $n$ of AdaIN in the DVG module is set to 6 in the paper.

\begin{figure*}[htp]
  \centering
  \includegraphics[width=1.0\textwidth]{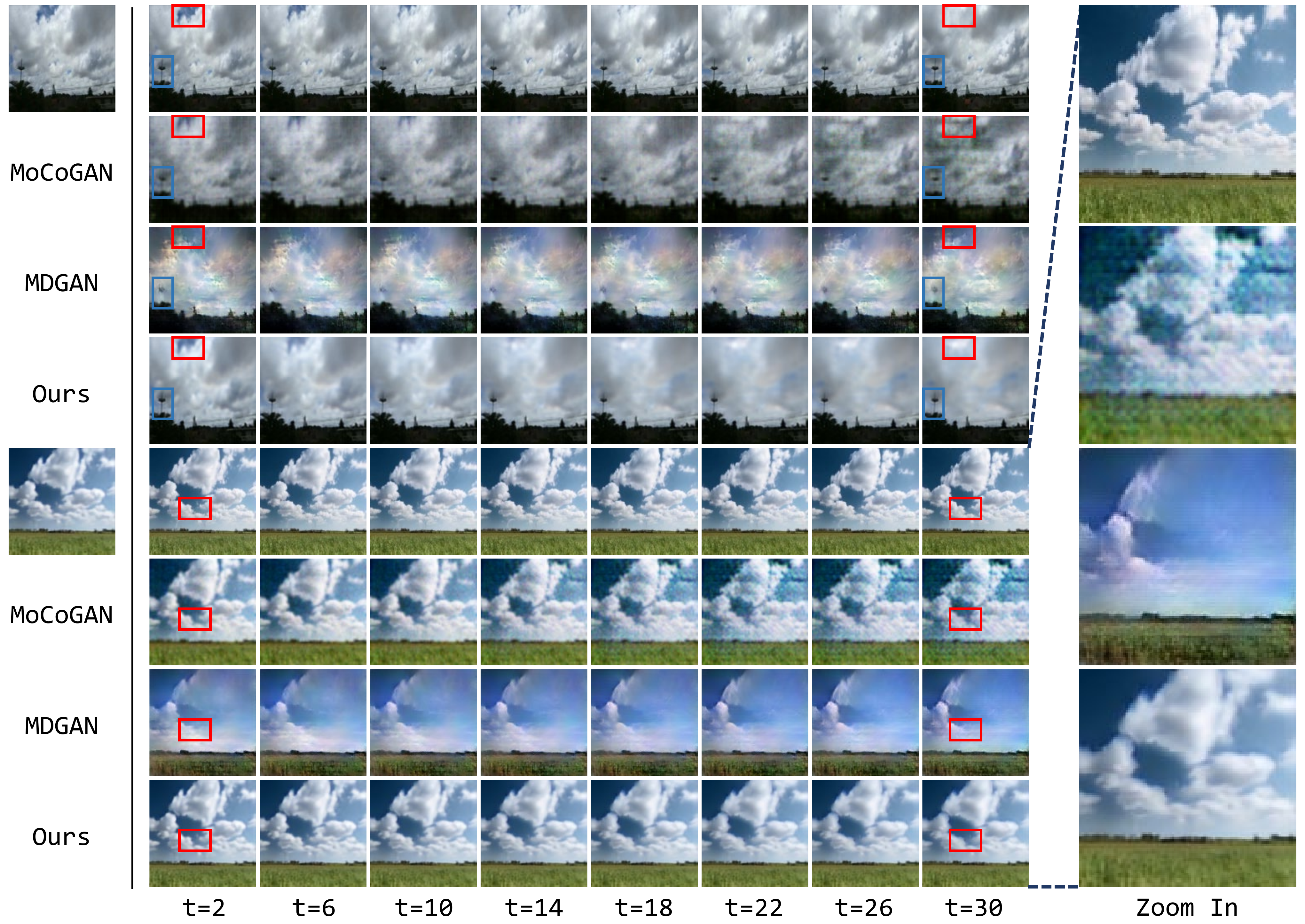}
  \caption{Generative results compared with MoCoGAN~\cite{Mocogan} and MDGAN~\cite{MDGAN} on the Sky Time-lapse dataset. The first column lists two different landscape images as the start frames, and the middle eight columns are generated video frames by different methods at different times. Right four enlarged images are long-term results at t=30 for a better visual comparison. We mark some dynamic and still details in red and blue rectangles, and please zoom in them for more precise comparison.}
  \label{fig:comparison2}
\end{figure*}

\section{Experiments}
In this section, many experiments are conducted to evaluate the method's effectiveness in various aspects of the Sky Time-lapse dataset. We first qualitatively and quantitatively compare our approach with two state-of-the-art method, \ie, MoCoGAN~\cite{Mocogan} and MDGAN~\cite{MDGAN}, and then conduct ablation studies to illustrate the effects of the structure and loss functions of our approach. Furthermore, we make a human study to demonstrate that our method can generate high-quality and dynamic video frames. Finally, we analyze the diversified generation ability of the proposed DTVNet.

\subsection{Datasets and Implementations Details}
\noindent\textbf{Sky Time-lapse} dataset~\cite{MDGAN} includes over 5000 time-lapse videos that are cut into short clips from Youtube, which contain dynamic sky scenes such as the cloudy sky with moving clouds and the starry sky with moving stars. Across the entire dataset, there are 35,392 training video clips and 2,815 testing video clips, each containing 32 frames. The original size of each frame is $3 \times 640 \times 360$, and we resize each frame into a square image of size ${3 \times 128 \times 128}$ as well as normalize the color values to [-1, 1].

\noindent\textbf{Beach} dataset~\cite{VGAN} contains 6,293 clips from 248 videos, which mainly consist of coastal beach scenes that have sky, cloud, water, sand, rocks, \etc. Each clip has 32 frames and a total of 201,376 frames. The original size of each frame is ${3 \times 128 \times 128}$, and we normalize the color values to [-1, 1].

\noindent\textbf{Evaluation Metric.}
We use \emph{Peak Signal-to-Noise Ratio} (PSNR) to evaluate the frame quality at the pixel level, and \emph{Structural Similarity} (SSIM)~\cite{SSIM} to measure the structural similarity between synthesized and natural video frames. However, these two metrics could not well evaluate the motion information of video sequences, so we introduce another metric, Flow-MSE~\cite{flow2}, to calculate the difference in the optical flow between generated video sequences and ground truth sequences. Furthermore, we conduct a \emph{Human Study} to evaluate the visual quality of generated videos by real persons.

\noindent\textbf{Implementation Details.}
We follow the training scheme described in~\ref{TS}. The OFE module inputs the flow sequence $\boldsymbol{U}_{1\sim T}$ ($\in\mathbb{R}^{2 \times32 \times 128 \times 128}$) and produces the normalized motion vector $\bar{\boldsymbol{f}}$ ($\in\mathbb{R}^{512}$). The DVG module inputs the start frame $\boldsymbol{I}_{0}$ ($\in\mathbb{R}^{3 \times 128 \times 128}$) and $\bar{\boldsymbol{f}}$ to generate the flow sequence ${\hat{\boldsymbol{U}}}_{1\sim T}$ ($\in\mathbb{R}^{2 \times32 \times 64 \times 64}$) and synthesized video frames ${\hat{\boldsymbol{I}}}_{1\sim T}$ ($\in\mathbb{R}^{3 \times32 \times 128 \times 128}$). During the training stage, we use Adam~\cite{adam} optimizer for all modules and set $\beta_{1}$=0.99, $\beta_{2}$=0.999. The initial learning rate is set to $3e^{-4}$, and it decays by ten every 150 epochs. We train the DTVNet for 200 epochs and the batch size is 12. We further test the inference speed of the DTVNet that can run 31 FPS with a single 1080 Ti GPU. Our model volume is 8.42M that is nearly a tenth of MDGAN~\cite{MDGAN}, \ie, 78.14M.

\begin{table}[b]
  \centering
  \caption{Metric evaluation results of MoCoGAN~\cite{Mocogan}, MDGAN~\cite{MDGAN}, and our approach on the Sky Time-lapse dataset. The up arrow indicates that the larger the value, the better the model performance, and vice versa.}
  \label{table:com2}
  \begin{tabular}{C{60pt}C{40pt}C{40pt}C{40pt}}
  \toprule
  \multirow{1}{*}{\textbf{Method}} & \multirow{1}{*}{\textbf{PSNR $\uparrow$}} & \multirow{1}{*}{\textbf{SSIM $\uparrow$}} & \multirow{1}{*}{\textbf{Flow-MSE $\downarrow$}} \\
  \midrule
  MoCoGAN~\cite{Mocogan} & 23.867 & 0.849 & 1.365\\
  MDGAN~\cite{MDGAN} & 23.042 & 0.822 & 1.406\\
  Ours & \textbf{29.917} & \textbf{0.916} & \textbf{1.275}\\
  \bottomrule
  \end{tabular}
\end{table}

\begin{table}[b]
  \centering
  \caption{Quantitative experimental results compared with MoCoGAN~\cite{Mocogan} and MDGAN~\cite{MDGAN} on the Beach dataset. The up arrow indicates that the larger the value, the better the model performance, and vice versa.}
  \label{table:com_beach}
  \begin{tabular}{C{60pt}C{40pt}C{40pt}C{40pt}}
  \toprule
  \multirow{1}{*}{\textbf{Method}} & \multirow{1}{*}{\textbf{PSNR $\uparrow$}} & \multirow{1}{*}{\textbf{SSIM $\uparrow$}} & \multirow{1}{*}{\textbf{Flow-MSE $\downarrow$}} \\
  \midrule
  MDGAN~\cite{MDGAN} & 16.195 & 0.802 & 1.046 \\
  MoCoGAN~\cite{Mocogan} & 21.413 & 0.826 & 0.822 \\
  Ours & \textbf{26.228} & \textbf{0.879} & \textbf{0.764} \\
  \bottomrule
  \end{tabular}
\end{table}

\subsection{Comparison with State-of-the-arts} \label{comparison_sotas}
\noindent\textbf{Qualitative Results.}
We conduct and discuss qualitative results, compared with MoCoGAN and MDGAN, on the Sky Time-Lapse dataset. As shown in Figure \ref{fig:comparison2}, we randomly sample two videos from the test dataset with different dynamic speeds (the cloud in the top half moves much faster than the bottom). The first column shows the start frames of two videos, while the second to ninth columns are generated video frames by different methods at different times. Note that the first and fifth rows are ground truth frames, and the test models of other methods are supplied by official codes.

Results show that our method can keep better content information than other SOTA methods (comparing the generated results in the column at a special time) and capture the dynamic motion (viewing the generated results in a row). In detail, the generated sequences produced by MoCoGAN (second and sixth rows) become more and more distorted over time. Thus the quality and motion can not be well-identified. The results produced by MDGAN suffer from the distortion in color and can not reasonably keep the content.
Specifically, we mark some dynamic and still details in red and blue rectangles. Results show that our method can well keep the content of the still objects while predicting reasonable dynamic details, which outperforms all other state-of-the-art methods. 

\noindent\textbf{Quantitative Results.}
We choose PSNR, SSIM, and Flow-MSE metrics to quantitatively evaluate the effectiveness of our proposed method on the Sky Time-lapse dataset. In detail, we use all start frames in the test dataset to generate corresponding videos by different methods and then calculate metric scores with ground truth videos.

As shown in Table~\ref{table:com2}, our approach gains +6.05 and +6.875 improvements for the PSNR metric as well as 0.067 and 0.094 for the SSIM metric compared to MoCoGAN and MDGAN, respectively. For the Flow-MSE metric, our method achieves the lowest value, which means that our generated video sequences are the closest to the ground truth videos in terms of motion.
Overall, evaluation results indicate that our approach outperforms the other two baselines in all three metrics, which illustrates that our model can generate more high-quality and dynamic videos than other methods.

Moreover, we conduct a comparison experiment with SOTAs on the Beach dataset. As shown in Table~\ref{table:com_beach}, our approach gains a significant improvement in PSNR metric than SOTA methods and improves 0.077 and 0.053 for the SSIM metric compared to MDGAN and MoCoGAN. For the Flow-MSE metric, our method achieves the lowest value, \ie, 0.764, which means that our generated video sequences are the closest to the ground truth videos in terms of the motion. Besides, we find that metrics on the Beach dataset are lower than that on the Sky Time-lapse dataset, which indicates that the Beach dataset is more challengeable.

\begin{table}[t]
  \centering
  \caption{Human study about frame and video quality evaluations on the Sky Time-lapse dataset.}
  \label{table:HS}
  \begin{tabular}{C{80pt}C{68pt}C{68pt}}
  \toprule
  \multirow{1}{*}{\textbf{Comparison Methods}} & \multirow{1}{*}{\textbf{Frame Quality Score}} & \multirow{1}{*}{\textbf{Video Quality Score}} \\
  \midrule
  MoCoGAN~\vs~GT & ~3~\vs~97 & ~2~\vs~98 \\
  MDGAN~\vs~GT & ~2~\vs~98 & ~1~\vs~99 \\
  Ours~\vs~GT & \textbf{12}~\vs~\textbf{88} & ~\textbf{7}~\vs~\textbf{93} \\
  \midrule
  Ours~\vs~MoCoGAN & \textbf{94}~\vs~6~ & \textbf{96}~\vs~4~ \\
  Ours~\vs~MDGAN & \textbf{97}~\vs~3~ & \textbf{99}~\vs~1~ \\
  \bottomrule
  \end{tabular}
  \end{table}

\begin{figure*}[htp]
  \centering
  \includegraphics[width=1.0\textwidth]{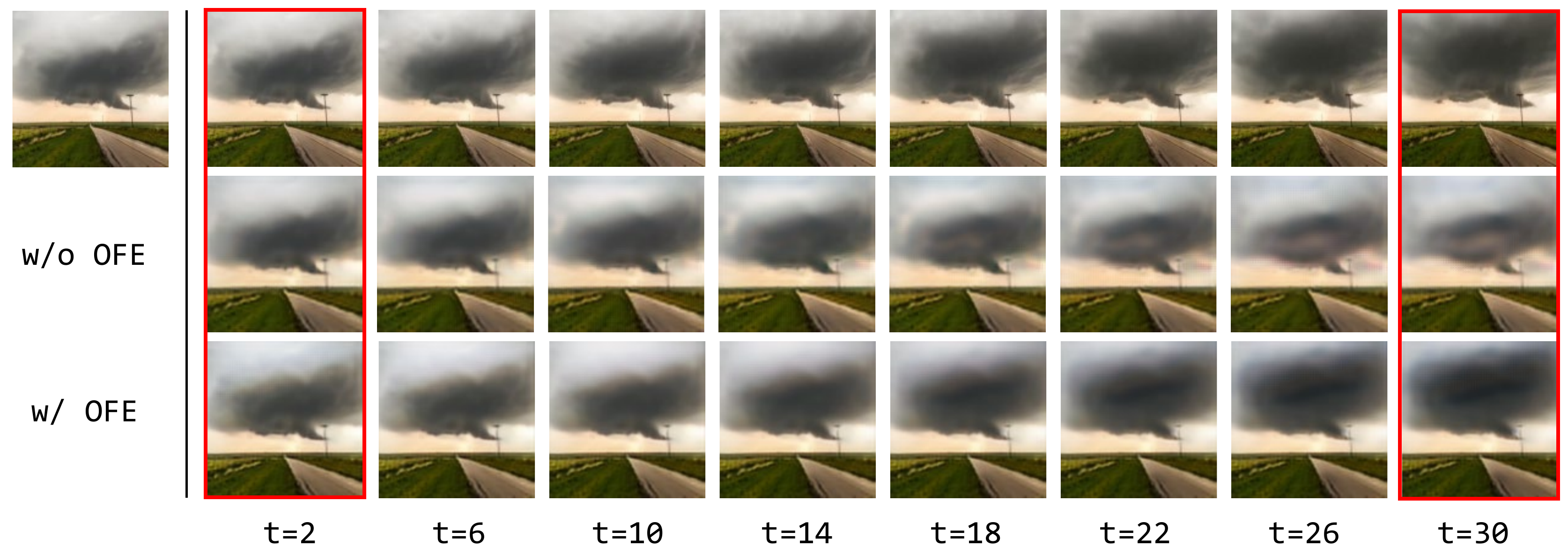}
  \caption{A toy experiment for testing the effect of the \emph{flow stream} of our approach. Images in the first row are ground truth frames at different times. The second and third rows are generated video frames without and with flow stream of our approach from the same start frame (the image in the first column). Please zoom in red rectangles for more precise comparison.}
  \label{fig:flow}
\end{figure*}

\begin{figure*}[t]
  \centering
  \includegraphics[width=1.0\textwidth]{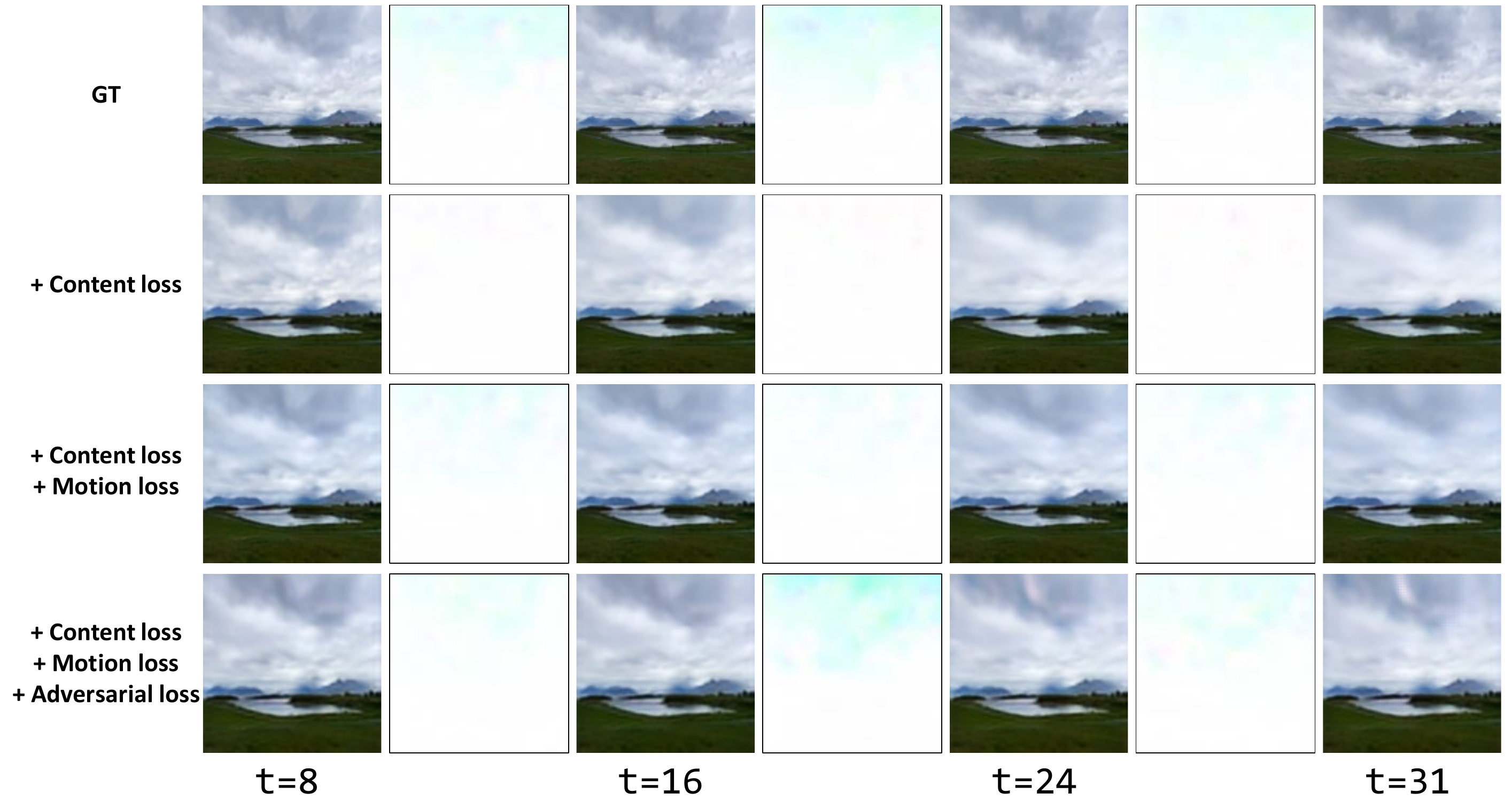}
  \caption{Qualitative comparisons of our approach with different loss terms on Sky Time-lapse dataset. GT indicates the ground truth of the video frames. The images in even columns are the visualization results of the optical flow between two adjacent images. Please zoom in for more details.}
  \label{fig:loss}
\end{figure*}

\noindent\textbf{Human Study.}
We further perform a human study for artificially evaluating the video frame quality. In detail, we first randomly choose 100 start images from the test dataset and generate related videos by different methods. Then the generated videos by two different methods are simultaneously shown to a real person, and we can get a result indicating "which one is better." The experiment mentioned above is conducted 30 times by 30 real workers, and we take the average of all results as the final result. Specifically, we split video quality evaluation into two aspects: the single-frame quality evaluation (selecting frame 16 for each video sequence) and the dynamic video evaluation, which evaluates content and motion information separately. Note that the image is shown for 2 seconds, and the video is played only once for a worker.

From the comparison results shown in Table~\ref{table:HS}, our method and other two baselines have a lower quality score than ground truth images, which means that there are still many challenges for the video generation task. Nevertheless, our method has a better quality score than others: 12\vs 3/2 in the frame and 7\vs 2/1 in the video.
To make a more intuitive comparison between our method and the others, we conduct another experiment in the bottom two rows. Results demonstrate that our approach outperforms the other two SOTA methods.

\subsection{Ablation Study}
In this section, we conduct several ablation studies on the Sky Time-Lapse dataset to analyze the contribution of the \emph{optical flow encoder} submodule and the effectiveness of different loss terms.

\begin{figure*}[htp]
  \centering
  \includegraphics[width=1.0\textwidth]{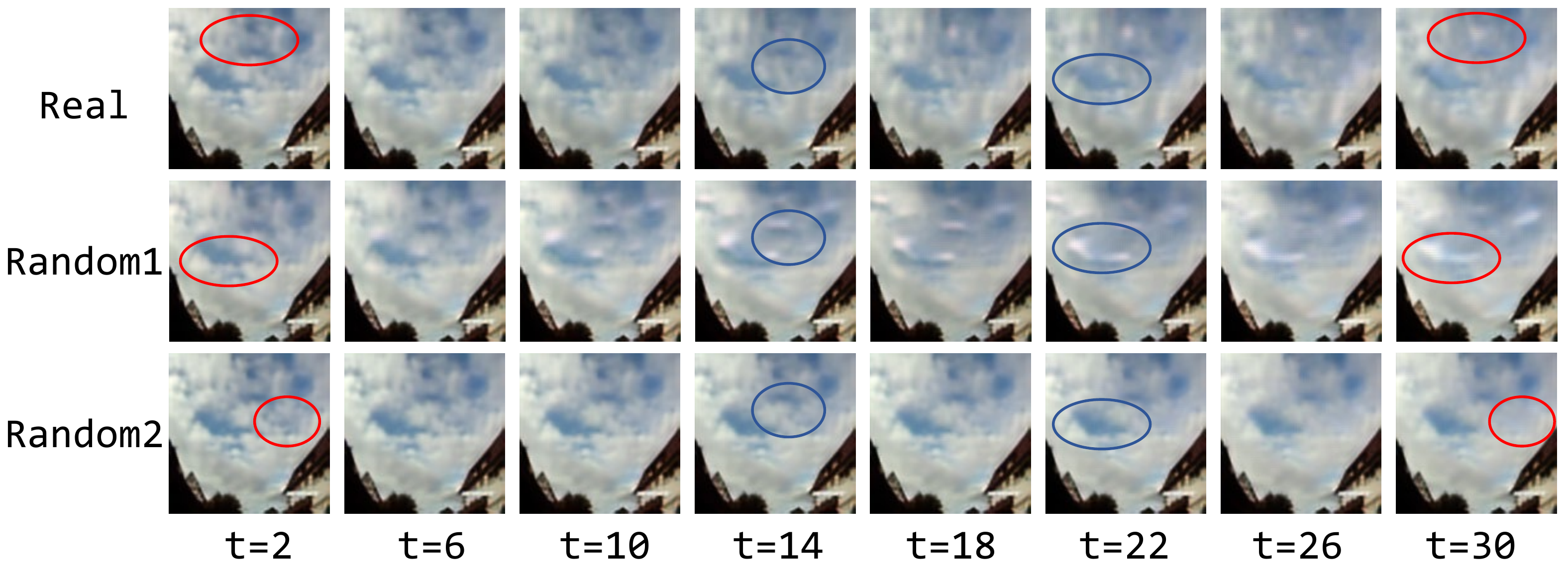}
  \caption{Diversified video generation experiment. Given the same start frame, our approach can generate different videos under the different motion vectors, \eg, encoded from the real video (denoted as \textit{Real}) and sampled from the normalization distribution (denoted as \textit{Random1} and \textit{Random2}). Please zoom in the red rectangles to compare each generated video in temporal and blue ellipses to compare different video frames generated by different motion vectors.}
  \label{fig:noise}
\end{figure*}

\noindent\textbf{Influence of OFE.} To evaluate the effectiveness of the optical flow encoder module, we conduct an ablation experiment that with or without the OFE module of our proposed DTVNet. As shown in Figure~\ref{fig:flow}, images in the first row are ground truth frames, and the second and third rows illustrate generation results without and with the OFE module, respectively.

By analyzing the results, we find that the generated frames lost the motion information and could not capture the precise movement and texture of the cloud without the OFE module (as shown in the second row of Figure~\ref{fig:flow}). In detail, the generated frames remain almost stationary and become more obscure as time goes on.
When adding the OFE module, the model can generate high-quality and long-term dynamic video (as shown in the third row of Figure~\ref{fig:flow}), which demonstrates that the OFE is critical for synthesizing photorealistic and dynamic videos. Please zoom in the red rectangles to compare long-term generation results of different structures.

\begin{table}
\centering
\caption{Quantitative comparisons of our approach with different loss terms on Sky Time-lapse dataset.}
\label{table:loss}
\begin{tabular}{C{60pt}C{40pt}C{40pt}C{46pt}}
\toprule
\textbf{Method} & \textbf{PSNR} $\uparrow$ &\textbf{SSIM} $\uparrow$ &\textbf{Flow-MSE} $\downarrow$\\
\midrule
$\mathcal{L}_{C}$ & 28.768 & 0.897 & 1.624\\
$\mathcal{L}_{C} + \mathcal{L}_{M}$ & 29.364 & 0.906 & 1.509\\
$\mathcal{L}_{C} + \mathcal{L}_{M} + \mathcal{L}_{adv}$ & \textbf{29.917} & \textbf{0.916} & \textbf{1.275}\\
\bottomrule
\end{tabular}
\end{table}

\noindent\textbf{Influence of Loss Functions.} To further illustrate the effectiveness of different loss functions, \ie, content loss, motion loss, and adversarial loss, we conduct qualitative and quantitative experiments with different loss functions on the Sky Time-lapse dataset. As shown in Figure~\ref{fig:loss}, the first row indicates the ground truth of the video frames, and the other three rows are generation results under the supervision of different loss functions. Generated video frames in odd columns indicate that \emph{motion loss} and \emph{adversarial loss} can greatly improve the model performance. The generated video frames have more clear details and higher quality when gradually adding loss terms. \ie, results in the third row are better than the second row, and results in the fourth row are better than the third row.

We further visualize the optical flow between two adjacent images in even columns, and the result shows that our approach can well learn the motion information when generating video frames.
Also, we quantitatively evaluate the effectiveness of different loss functions and obtain a similar result. As shown in Table~\ref{table:loss}, the model obtains the best metric results, \ie, PSNR=29.917, SSIM=0.916, and Flow-MSE=1.275, when all the three loss functions are used.

\subsection{Diversified Video Generation}
Diversity is a critical factor in content creation. Besides generating a fixed target video from a single start frame, diversified video generation from one still landscape image is critical for practical application. In this section, we conduct an additional diversified video generation experiment to illustrate the advantage of our approach for using \emph{normalized motion vector} to provide motion information. 

In detail, different video frames are generated by DTVNet under the different motion vectors, \eg encoded from the real video (denoted as \textit{Real}) and sampled from the normalization distribution (denoted as \textit{Random1} and \textit{Random2}), as shown in Figure~\ref{fig:noise}. We can observe that our approach can generate diversified video frames by different motion vectors, which have the same content information but different motion information.

\begin{figure*}
  \centering
  \includegraphics[width=1.0\textwidth]{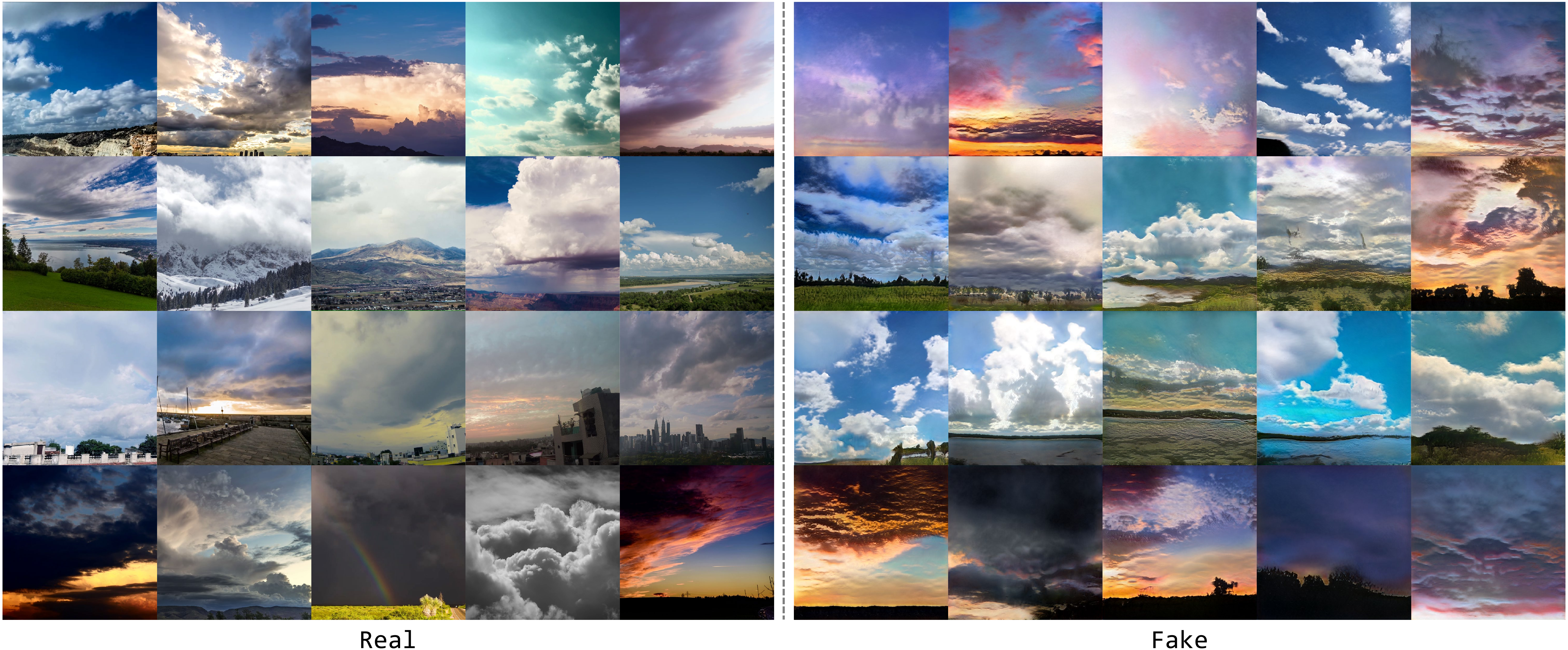}
  \caption{Real images (left) and generated images of StyleGAN2 (right) at $512 \times 512$ resolution on the proposed QST dataset. Please zoom in for more details.}
  \label{fig:stylegan}
\end{figure*}

\begin{table}[tp]
  \centering
  \caption{Comparison of different datasets.}
  \label{table:dataset}
  \begin{tabular}{C{70pt}C{30pt}C{30pt}C{60pt}}
  \toprule
  \multirow{2}{*}{\textbf{Datasets}} & \multirow{2}{*}{Clips} & \multirow{2}{*}{Frames} & \multirow{2}{*}{~~~Resolution} \\
  & & & \\
  \midrule
  Sky Time-lapse~\cite{MDGAN} & 38,207 & 1,222,624 & $~~~128 \times 128$\\
  Beach~\cite{VGAN} & 6,293 & 201,376 & $~~~128 \times 128$\\
  Quick-Sky-Time & 1,167 & 285,446 & $\geq 1,024 \times 1,024$ \\
  \bottomrule
  \end{tabular}
\end{table}

\section{Quick-Sky-Time Dataset}
With the improvement of computation and the demand for high-resolution image generation in practical applications, the current datasets~\cite{MDGAN,VGAN} are not enough to support the subsequent research on time-lapse video generation algorithms. Because these datasets contain images of low resolution and a limited quantity, thus we meticulously cut out 1,167 video clips in 216 time-lapse 4K videos collected from YouTube to form the high-quality Quick-Sky-Time (QST) dataset, which mainly consists of sky time-lapse scenes that have sky, cloud, mountain, house, \etc. Specifically, each short clip contains multiple frames (from a minimum of 58 frames to a maximum of 1,200 frames, a total of 285,446 frames), and the resolution of each frame is more than $1,024 \times 1,024$. We split the selected video clips into a training set (containing 1000 clips, totally 244,930 frames), a validation set (containing 100 clips, totally 23,200 frames), and a testing set (containing 67 clips, totally 17,316 frames). The specific comparison with some datasets is shown in Table~\ref{table:dataset}.

Our proposed QST dataset contains a variety of challenging scenes, such as different periods (daytime and nightfall), different scene objects (mountains, houses, \etc.), complex backgrounds, as well as light and shadow shifts, as shown in the left part of Figure~\ref{fig:stylegan}. At the same time, the QST dataset can be used for a variety of tasks, such as (high-resolution) video generation, (high-resolution) video prediction, (high-resolution) image generation, texture generation, image inpainting, \etc. In the following sections, we conduct a generation experiment to evaluate the difficulty of the dataset and use it for the video prediction task. 

\subsection{Generation on QST}
With the improvement of computation and further research of generative algorithms, some GAN-based approaches~\cite{stylegan,stylegan2,msggan} can generate high-resolution and high-quality images. In this section, the verified StyleGAN2~\cite{stylegan2} is used to conduct random generation experiments on the proposed QST dataset to prove that our QST dataset has a stable modality, \ie, different scenes in QST can be mapped by the noise of the same distribution. As viewed in the right part of Figure~\ref{fig:stylegan}, we show some generated scene images. Results show that a variety of images can be generated when training on the QST dataset, demonstrating the effectiveness and richness of the proposed QST dataset. However, some of the semantic details of the generated images exist in an artifact, such as house and meadow, which illustrates that our dataset is more challenging for the generation task. Also, we employ \emph{Fr\'echet Inception Distance} (FID)~\cite{FID} metric to evaluate image quality at the semantic level, and it obtains a relatively large score of 36.889 (the smaller, the better), which again demonstrates the difficulty of the dataset.

\begin{figure*}[ht]
  \centering
  \includegraphics[width=1.0\textwidth]{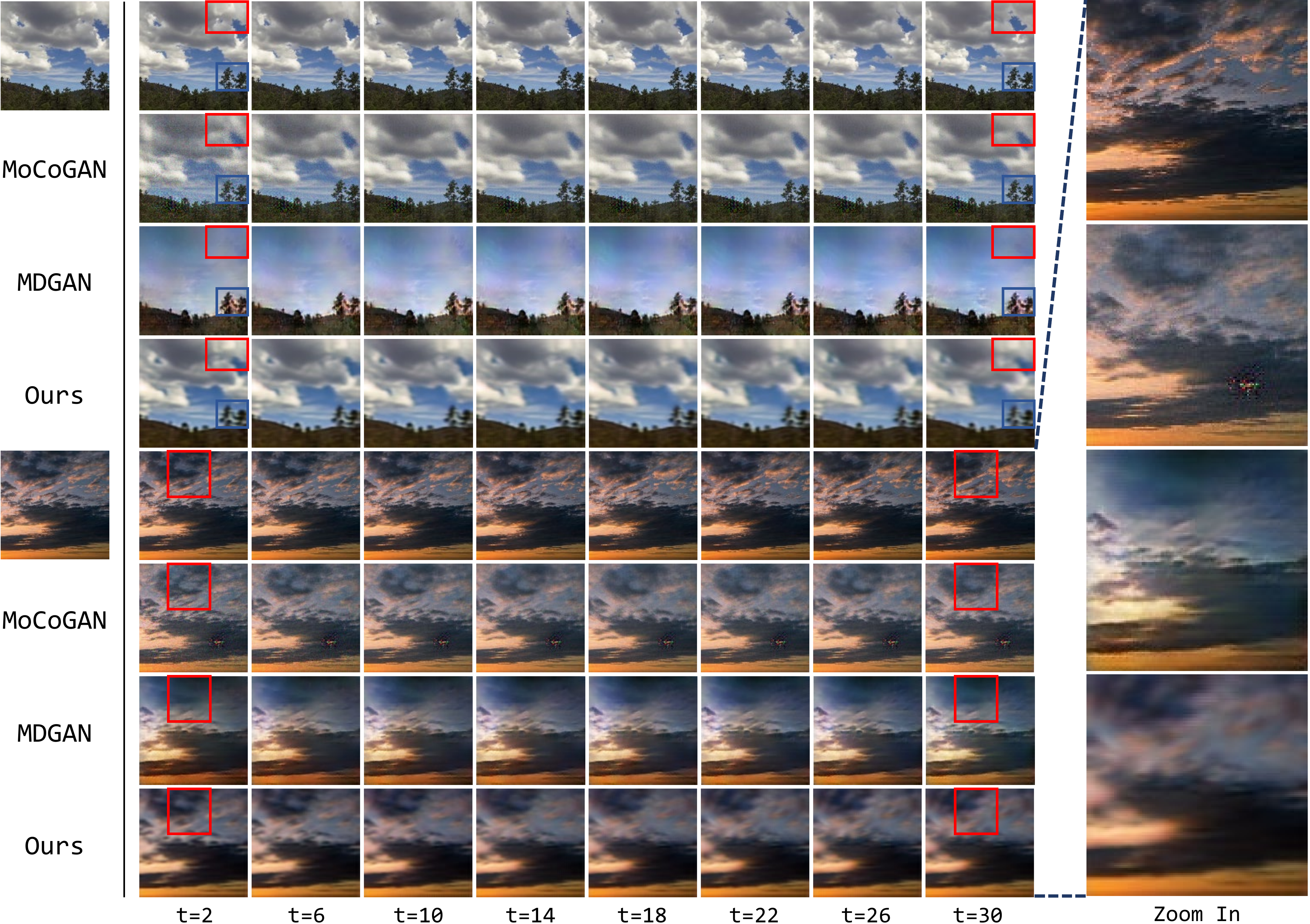}
  \caption{Qualitative experimental results compared with MoCoGAN~\cite{Mocogan} and MDGAN~\cite{MDGAN} on the proposed QST dataset. The first column lists two different landscape images as the start frames, and the middle eight columns are generated video frames by different methods at different times. Right four enlarged images are long-term results at t=30 for a better visual comparison. We mark some dynamic and still details in red and blue rectangles, and please zoom in on them for a more precise comparison.}
  \label{fig:com_qst}
\end{figure*}

\begin{table}[tp]
  \centering
  \caption{Quantitative experimental results compared with MoCoGAN~\cite{Mocogan} and MDGAN~\cite{MDGAN} on the proposed QST dataset. The up arrow indicates that the larger the value, the better the model performance, and vice versa.}
  \label{table:com_qst}
  \begin{tabular}{C{60pt}C{40pt}C{40pt}C{40pt}}
  \toprule
  \multirow{1}{*}{\textbf{Method}} & \multirow{1}{*}{\textbf{PSNR $\uparrow$}} & \multirow{1}{*}{\textbf{SSIM $\uparrow$}} & \multirow{1}{*}{\textbf{Flow-MSE $\downarrow$}} \\
  \midrule
  MDGAN~\cite{MDGAN} & 20.791 & 0.783 & 1.470 \\
  MoCoGAN~\cite{Mocogan} & 23.565 & 0.824 & 1.406 \\
  Ours & \textbf{27.544} & \textbf{0.875} & \textbf{1.332} \\
  \bottomrule
  \end{tabular}
\end{table}

\subsection{Video Prediction on QST}
We also conduct qualitative and quantitative experiments with SOTA methods on the new QST dataset using the same experimental settings as Section~\ref{comparison_sotas}. As shown in Figure~\ref{fig:com_qst}, we present qualitative experimental results of our method with different SOTA methods (at $1,024 \times 1,024$ resolution). MoCoGAN suffers from the pattern collapse during training, where it almost generates identical images for each frame without any motion information, while MDGAN can not well keep the semantic information of the input landscape image. Our method can maintain image semantics while achieving better motion information, consistent with the previous experimental results on other datasets.  As shown in Table~\ref{table:com_qst}, we further conduct a quantitative comparison with SOTA methods. Our method achieves the best scores on all metrics, \ie, 27.544 on PSNR, 0.875 on SSIM, and 1.332 on Flow-MSE, demonstrating the superiority of our method for generating higher quality images and better encoding motion information than other methods. However, above qualitative experimental results are not satisfactory, and there is still a large room for improvement in the QST dataset. In the future, we will further explore high-quality and high-resolution video generation.

\section{Conclusions}
This paper proposes a novel end-to-end one-stage dynamic time-lapse video generation framework, \ie, DTVNet, to generate diversified time-lapse videos from a single landscape image. The \emph{Optical Flow Encoder} submodule maps a sequence of optical flow maps to a \emph{normalized motion vector} that encodes the motion information inside the generated video. The \emph{Dynamic Video Generator} submodule contains motion and content streams that learn the movement and texture of the generated video separately, as well as an encoder and a decoder to learn shared content features and construct target video frames, respectively. During the training stage, we design three loss functions, \ie, content loss, motion loss, and adversarial loss, to the network, in order to generate high-quality and diversified dynamic videos. We exclude the OFE module and directly sample from normalization distribution as the motion vector during the testing stage, which reduces the network computing overhead and supply diversified motion information simultaneously. Furthermore, extensive experiments demonstrate that our approach is capable of generating high-quality and diversified videos. Also, we propose a large-scale and high-resolution Quick-Sky-Time dataset to further evaluate different approaches, which can be viewed as a new benchmark for high-quality video and image generation tasks.

We hope our study will help researchers and users achieve more effective works in the video generation task. We will explore how to efficiently produce higher-resolution and higher-quality videos in the future.

\bibliographystyle{IEEEtran}
\bibliography{IEEEexample}

\end{document}